\documentclass{article}
\usepackage{spconf,amsmath,graphicx}
\usepackage{amssymb,url}

\title{Medical Waste Sorting: a computer vision approach for assisted primary sorting}
\name{A. Bruno$^{1}$, C.Caudai$^{1}$, G.R. Leone$^{1}$, M. Martinelli$^{1}$, D. Moroni$^{1}$, F. Crotti$^{2}$\thanks{Correspondence should be addressed to: massimo.martinelli@isti.cnr.it}}
\address{$^{1}$ Institute of Information Science and Technologies, National Research Council, Pisa, Italy\\ $^{2}$ CISA, Lucca, Italy}
\begin{document}
%
\maketitle
\begin{abstract}
Medical waste, i.e. waste produced during medical activities in hospitals, clinics and laboratories, represents hazardous waste whose management involves special care and high costs. However, this kind of waste contains a significant fraction of highly valued materials that can enter a circular economy process. To this end, in this paper, we propose a computer vision approach for assisting in the primary sorting of medical waste. The feasibility of our approach is demonstrated on a representative dataset we collected and made available to the community, with which we have trained a model that achieves 100\% accuracy, and a new dataset on which the trained model exhibits good generalization.


\end{abstract}

\section{Introduction}
\label{sec:introduction}
Medical Waste (MW) is a growing concern as healthcare facilities produce a large amount of hazardous waste daily. 
Such waste requires special handling and treatment that imposes dedicated management approaches. Generally, medical waste is disposed into special containers in the place where it is produced. Such containers are then sealed and should be disposed of within a rather short timeframe. Disposal (mainly through incineration or waste-to-energy) should be done in special treatment facilities, which are often located far away from where the waste has been produced. In addition, after disposal in the container, manual manipulation is not permissible due to the possible biological hazard. Humans can intervene in waste only in critical situations and should be equipped with special personal protection equipment.
According to various sources, the production of medical waste has been increasing in recent years due to factors such as the growth of the healthcare industry and the increasing use of disposable medical equipment \cite{cheng2009medical}; also, the outbreak of COVID-19 -- which has lead to high peaks in the production of possibly dangerous and infectious waste --  urged the creation of more scalable and sustainable ways to treat medical waste \cite{tsai2021analysis}. 
It should be observed that medical devices are made in many situations with first-grade virgin materials, especially polymers; moreover, there are significant fractions of glasses, textiles and metals. Therefore, sorting medical waste for recycling is a way to enable more sustainable waste management. However, such an operation should be done as the primary sorting, since it is impossible to manually intervene at a later stage due to the aforementioned hazards. Indeed, a secondary sorting would be possible only after medical waste's infectious and hazardous nature has been neutralized, e.g. by sterilization. Notice that, the manual primary sorting of medical waste is a labour-intensive and time-consuming process that might excessively increment the burden of operators in healthcare facilities, with a high potential for human error. To this end, a system for assisting the specialized staff during sorting, facilitating the procedure and reducing possible errors should be devised.
In this paper, we propose a research in this direction by introducing an artificial intelligence (AI) model based on deep learning and computer vision methods that can classify several classes of items composing standard medical waste. Based on our previous experience \cite{bruno2022efficient,bruno2022exploring}, we propose a convolutional neural network based on the EfficientNet family \cite{tan2019EfficientNet}. To train the network, a special dataset has been collected mimicking a waste collection table endowed with a stereo camera. The dataset is still growing but, based on the current data, we were able to train a network architecture achieving encouraging results. Differently from other studies on MW in the literature, the dataset is made publically available to foster further research \cite{dataset}.
The paper is organized as follows; in Section \ref{sec:related} we briefly survey related works, while in Section \ref{sec:materialsandmethods}, we introduce the contributed datasets and the network architecture. Current results are reported in Section \ref{sec:exp}, while Section \ref{sec:conclusion} ends the paper with ideas for future work.

\section{Related works}
\label{sec:related}
In recent years, computer vision technology has emerged as a promising solution to improve the accuracy and efficiency of medical waste sorting. Computer vision involves using computer algorithms to analyze and understand images or videos. By applying computer vision techniques to the sorting of medical waste, it is possible to automate the process, reducing the risk of human error and increasing the speed and accuracy of the sorting.
The use of machine learning and, more recently, of deep learning has been established as a promising solution for automating the medical waste sorting process.
A seminal work related to general waste is represented by the Trashnet dataset \cite{yang2016classification}; this relatively simple, nevertheless public dataset has triggered much research in image classification for waste sorting, becoming the \emph{de facto} benchmark on the subject, used for testing Convolutional Neural Network (CNN) architectures \cite{ozkaya2019fine} as well as the fusion of deep features \cite{ahmad2020intelligent}.  Images, taken using mobile devices, represent standard classes such as glass, paper, cardboard, plastic, metal and garbage. 

Moving to the specific class of medical waste, the interest seems to be more recent. In \cite{bian2021medical}, for instance, the authors used a two-step approach. First, they identify the object of interest in an image by resorting to standard image processing procedures based on background subtraction. Then, they use a CNN consisting of an optimized version of SSD-MobileNet \cite{liu2016ssd} to identify medical waste. Their system is able to classify syringes, hemostatic forceps, infusion bags and gloves with a recognition accuracy of more than 98.5\% and an average recognition speed of 52 milliseconds, making it suitable for real-time analysis and prompt feedback to operators.

In \cite{mythili2022concatenation}, the authors focus on the classification of medical waste with images acquired with complex backgrounds. To deal with this issue, they introduce a deep network with segmentation capabilities, named EnSegNet, and they concatenate a set of deep features with additional handcrafted features extracted from the segmented images (e.g. textural features  based on Gray-Level Co-occurrence Matrix (GLCM), Multi-level Local Binary Pattern (MLBP), Local Derivative Pattern (LDP) and Local Ternary Pattern (LTP)). Concatenaed features are then classified by a fully connected layer. Their
hybrid feature vector has a stronger discriminant ability compared to the single feature vector and the overall pipeline shows  that the  system attains a 93.7\% of classification accuracy for 100 trash
images. The dataset in any case seems to be small and not sufficiently representative. A more comprehensive work is presented in \cite{zhou2022deep}, where they applied  a fine-tuned version of ResNext \cite{xie2017aggregated} on a dataset of 3480 images and succeeded in correctly identifying eight kinds of medical waste with an accuracy of 97.2\%. Unfortunately, the dataset has not been disclosed.

After the COVID-19 outbreak, several implications for medical waste management emerged. For instance, possible infectious waste has been produced also in houses and in facilities where there is not usually such a production. This urged the creation of vision-based systems to detect different streams of waste. For instance, in \cite{kumar2021artificial}, they proposed a  waste type classification  based on the image-texture-dependent features, which provided an accurate sorting and classification before the recycling process starts. The classification model based on the support vector machine (SVM) classifier performs best (with 96.5\% accuracy, 95.3\% sensitivity, and 95.9\% specificity) in classifying waste types in the context of circular manufacturing and exhibiting the ability to manage COVID-related medical waste mixed.

In addition to the use of computer vision, some researchers have also explored the use of alternative technologies, such as infrared and hyperspectral imaging \cite{calvini2018developmentof} in waste sorting. Indeed, in such a way, it is possible to achieve a \emph{chemical imaging} and sort the waste depending on its material composition. This might turn very useful in medical waste sorting, separating plastic materials made of different kinds of polymers. While these technologies have shown promising results, they also face challenges, such as the high equipment cost and the need for specialized training make them unsuitable at present, at least in primary sorting.

\section{Materials and methods}
\label{sec:materialsandmethods}
Primary sorting of medical waste refers to the initial stage of the waste management process, where waste is separated into different categories based on its composition and properties. The primary sorting process is crucial for properly disposing of waste and for reducing waste-related environmental and health hazards. To provide effective support in this process, we have schematized a use case in which operators convey waste to a special disposal station consisting of a flat surface where they deposit garbage. Such a surface is under active surveillance by an imaging system which is able to classify acquired images into a set of predefined classes and provides feedback to operators and/or sends the waste to the right disposal/recycling pathway. 
Based on such a use case, we have collected a representative dataset (Section \ref{sec:dataset}) and designed a CNN architecture (Section \ref{sec:cnn}) for solving the computer vision task.




\subsection{Dataset acquisition}
\label{sec:dataset}

\begin{figure}[htp]
    \centering
    \includegraphics[width=7cm]{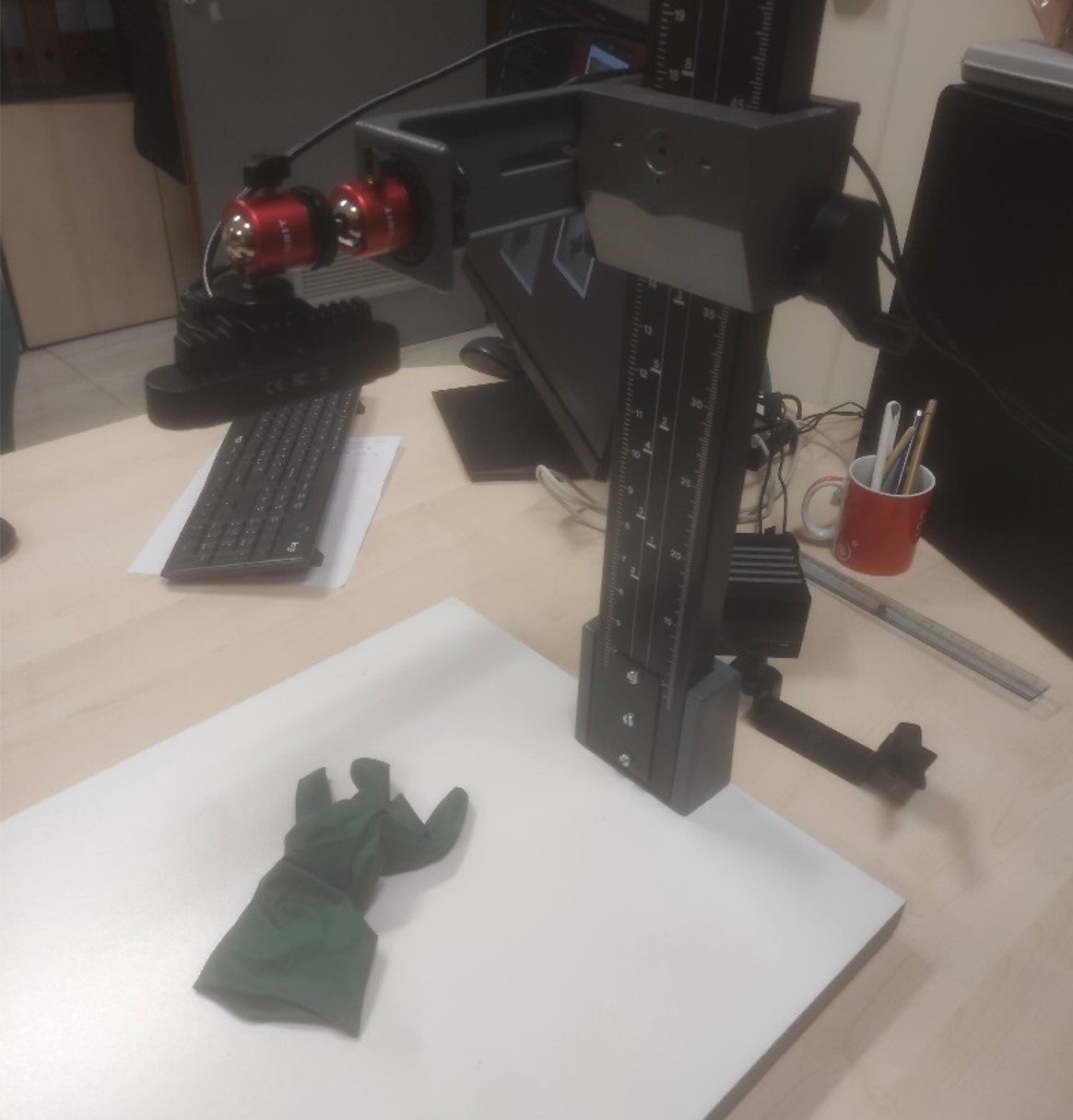}
    \caption{Image acquisition setup.}
    \label{fig:stand}
\end{figure}

\begin{figure}[htp]
    \centering
    \includegraphics[width=8cm]{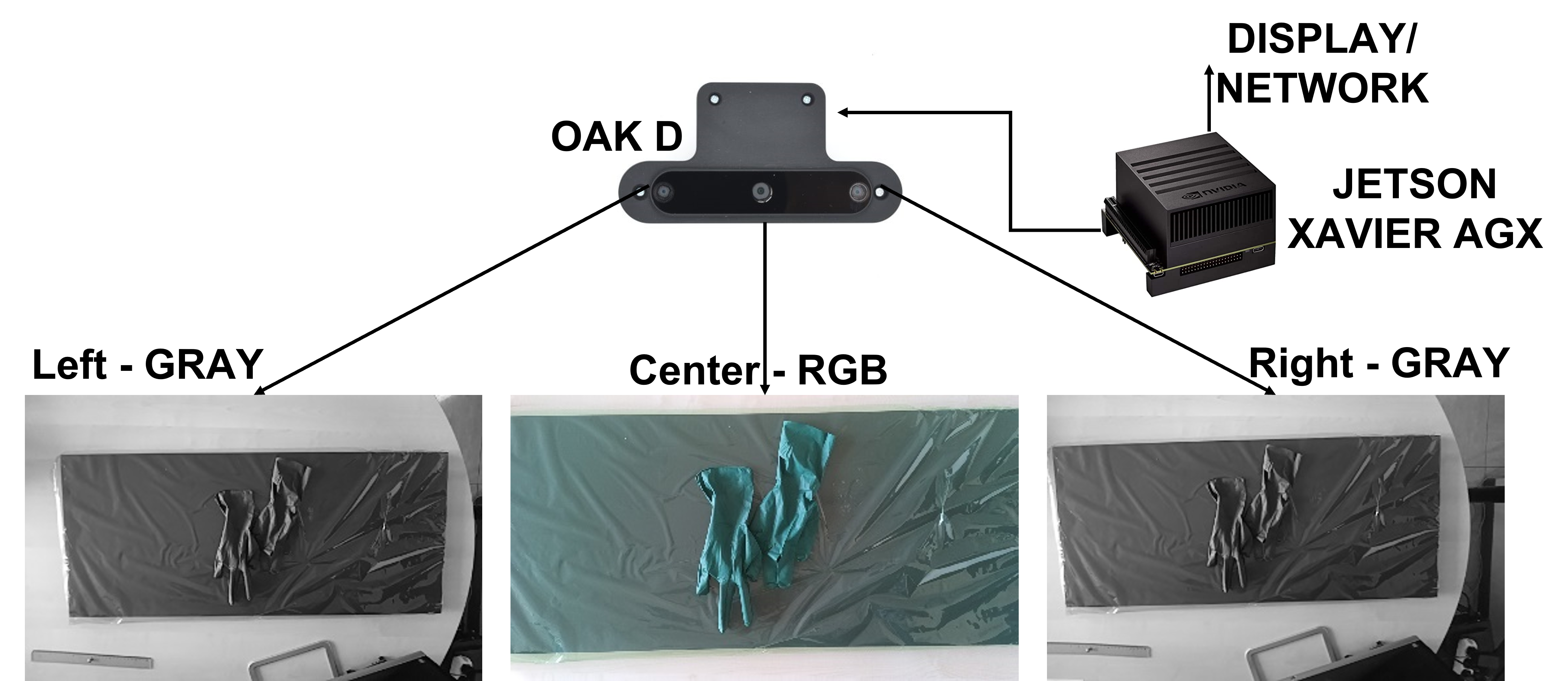}
    \caption{Overview of the acquisition using OAK-D camera \cite{OAK}.}
    \label{fig:setupexp}
\end{figure}

Medical waste was collected using a made-by-us lab table with an adjustable camera stand depicted in Figure \ref{fig:stand}.
As a camera device, we opted for a high‑resolution camera with depth vision and on‑chip machine learning powered by an Intel Movidius-X chip. Specifically, we used a Luxonis OAK-D device \cite{OAK}, equipped with an Intel Movidius Myriad-X VPU, USB 3.1, gen2 10gps, 2 x global shutter mono cameras (1Mp - 1280 x 800px) Omnivision OV9282 for stereo vision and 1x centre RGB camera (12MP - 4056 x 3040px) Sony IMX378. Considering the standard input layer size of the state-of-the-art convolution neural network, the full resolution achievable by the OAK-D camera was not used. Instead, each sample is composed of a triplet of images with the following characteristics: 1 RGB image of size 1920 x 1080 and 2 grayscale of size 640 x 400, from which depth information can be inferred. All the images can be aligned using the factory calibration of the OAK-D to compute depth and produce RGBD images. The inclusion of 3D information might be useful for refined spatial AI algorithms and for estimating the volume of waste at the moment of first disposal. In addition, in principle, some AI algorithms can be executed directly on the OAK-D device, fully embracing an edge computing perspective. The acquisition setup was completed by an NVIDIA Jetson Xavier AGX connected to the OAK-D camera for executing CNN inference and/or sending data and results through the network as depicted in Figure \ref{fig:setupexp}.
Variable natural and artificial illumination was used during the acquisitions which took place in two different rooms located at the Area delle Ricerca CNR in Pisa and CISA premises in Lucca, Italy.

\begin{figure*}
\begin{centering} 
\begin{tabular}{c c c c}
    \includegraphics[height=2.3cm]{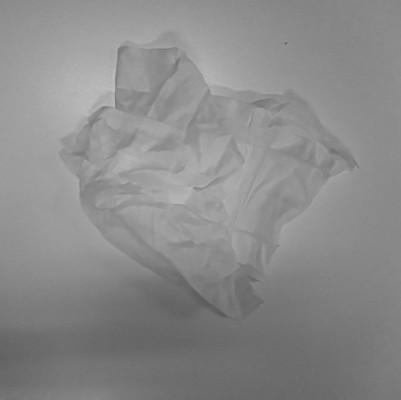} &
    \includegraphics[height=2.3cm]{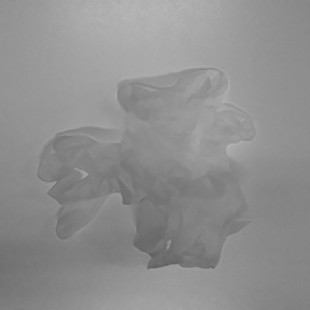} &
    \includegraphics[height=2.3cm]{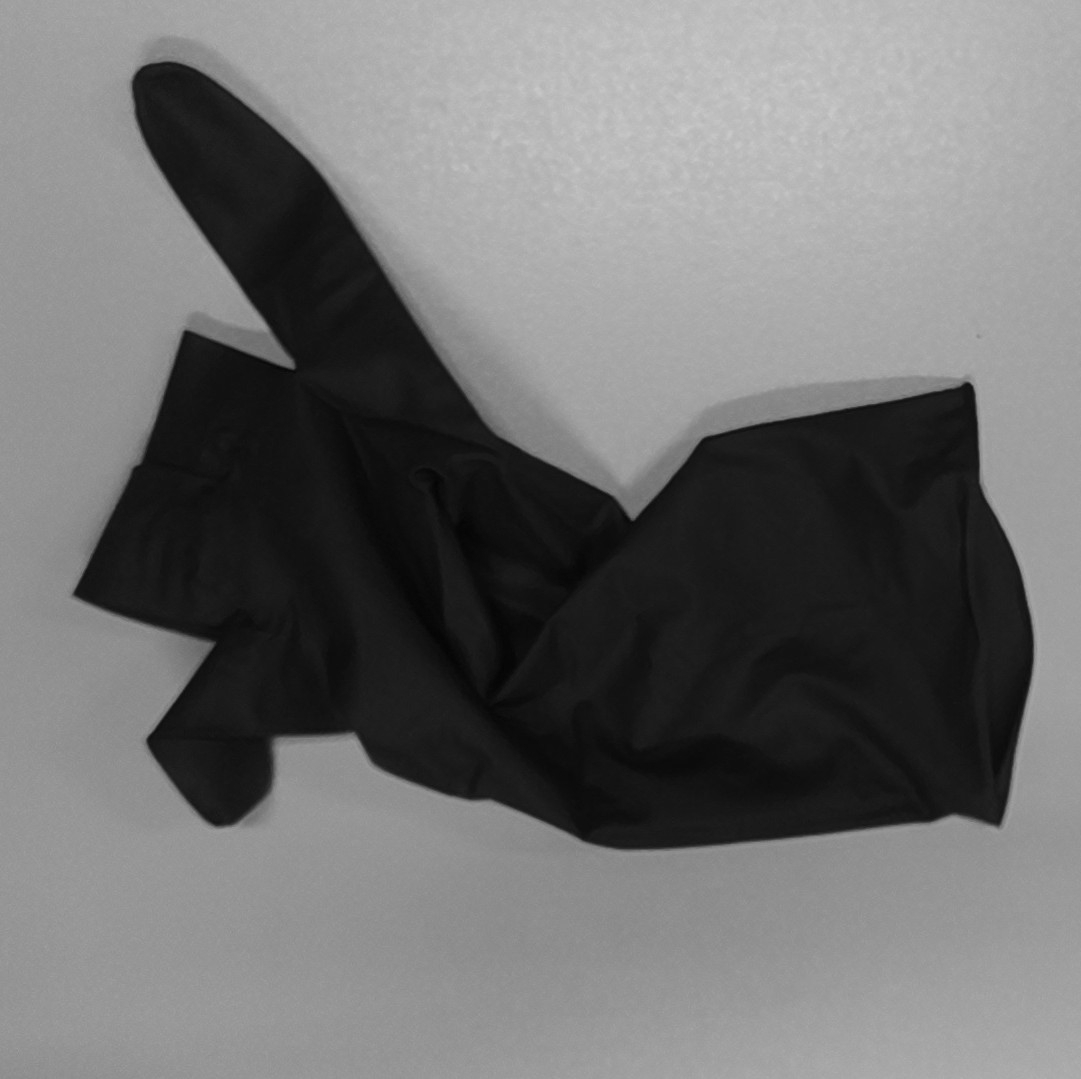} &
    \includegraphics[height=2.3cm]{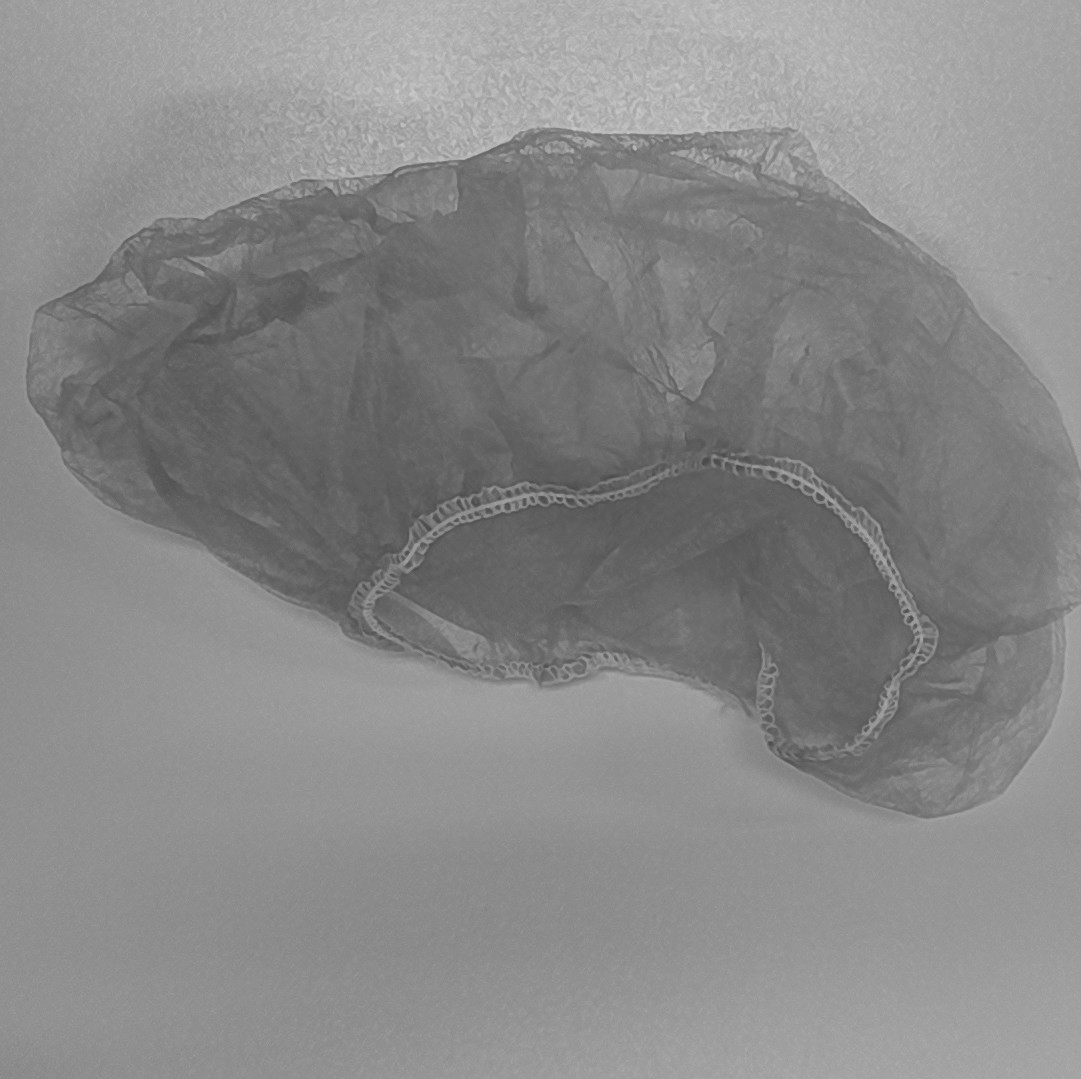}\\
    gauze & latex gloves & surgical gloves & medical cap \\
    \includegraphics[height=2.3cm]{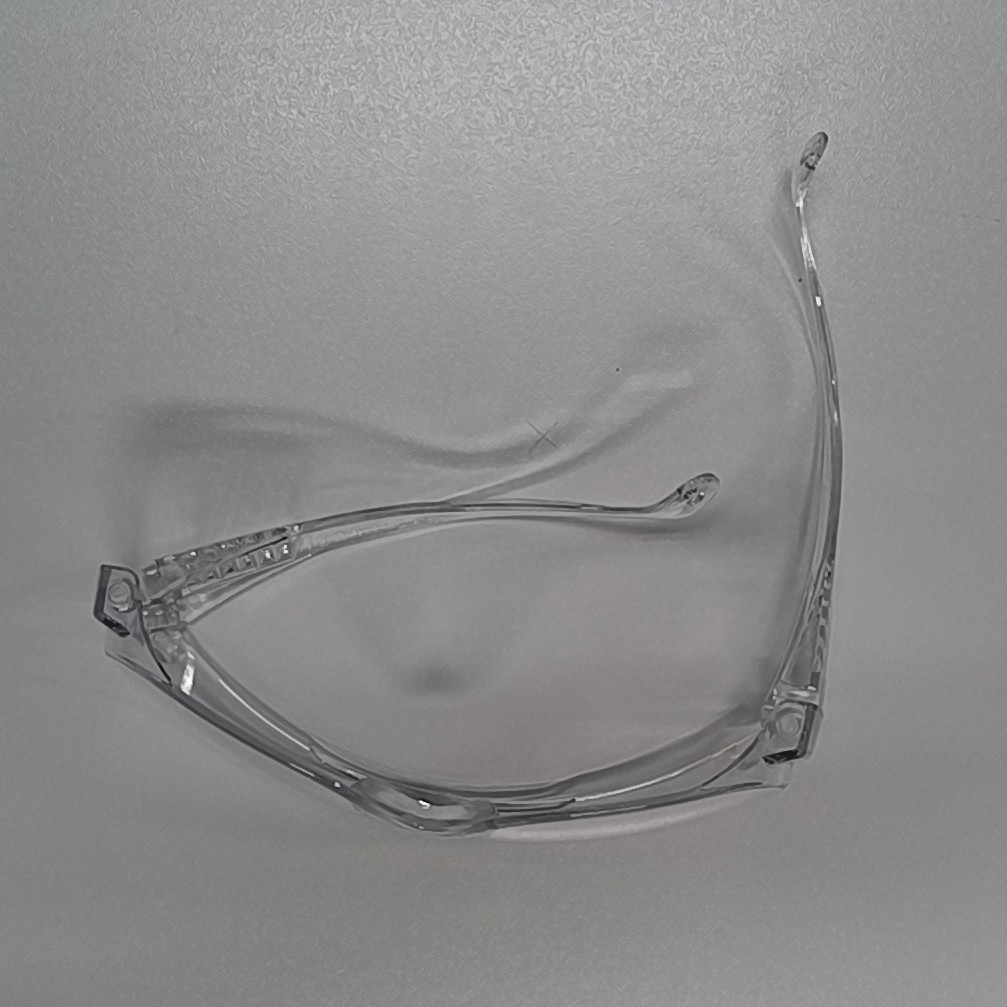} &
    \includegraphics[height=2.3cm]{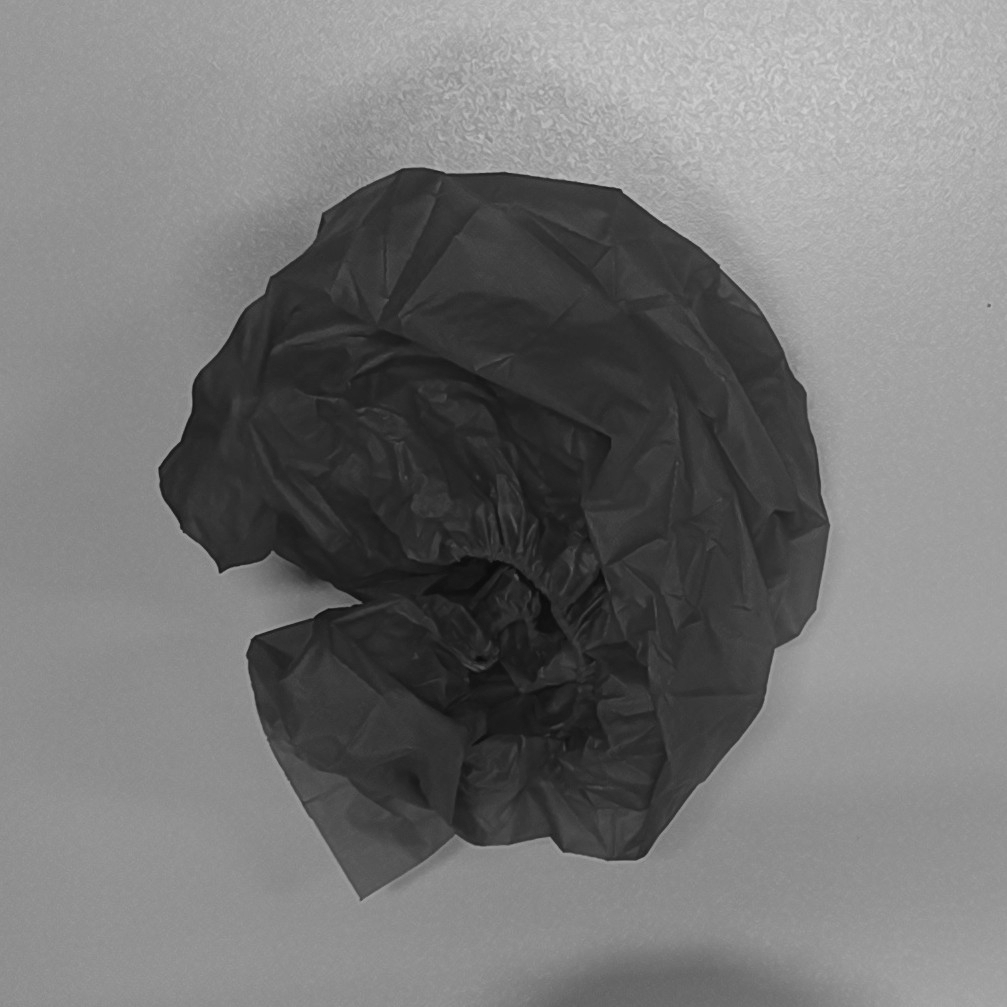} &
    \includegraphics[height=2.3cm]{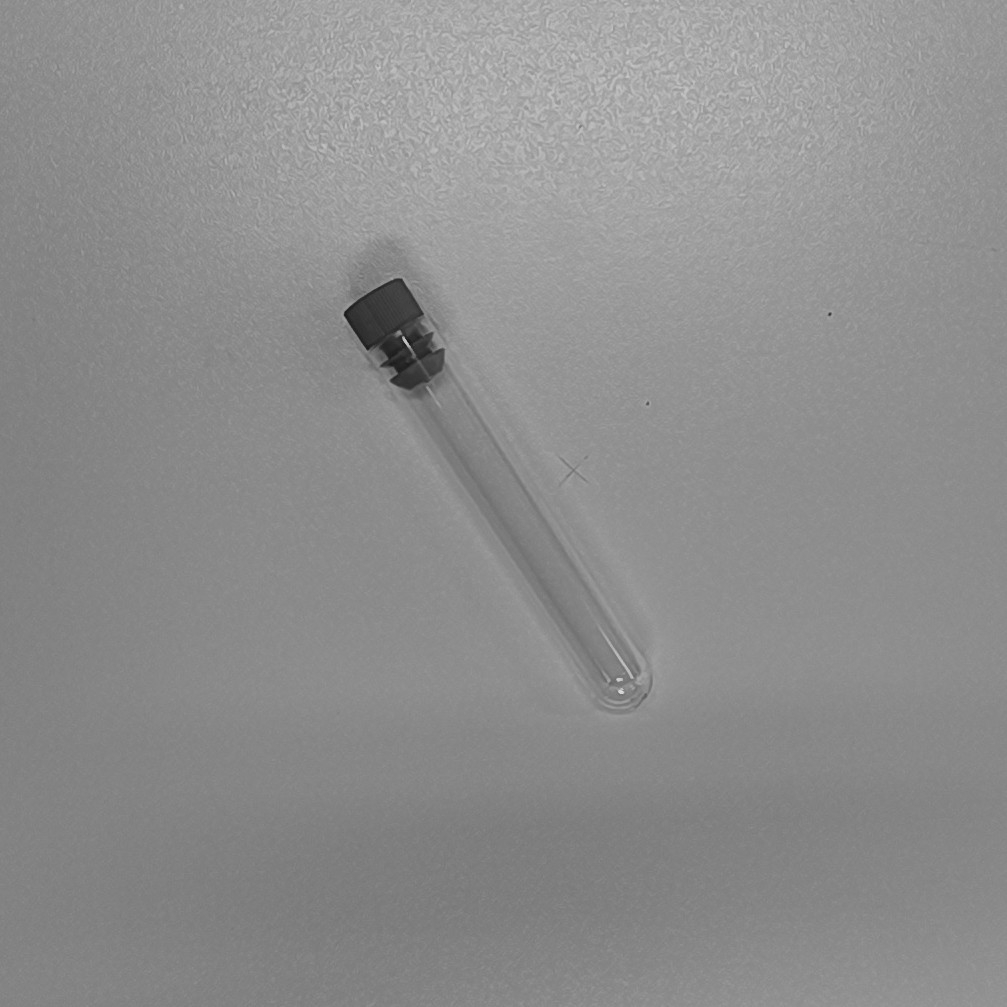} &
    \includegraphics[height=2.3cm]{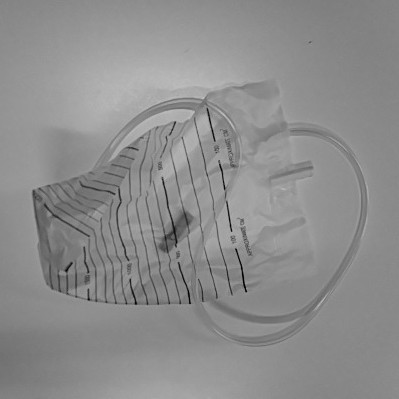} \\
    medical glasses & shoes & test tube & urine bag
    \end{tabular} 
    \caption{\label{fig:dataset}Examples from the classes of the dataset; for the class \texttt{gloves} we reported two different typologies. }
\end{centering}
\end{figure*}

Due to safety issues, medical waste was simulated using brand-new medical devices after a screening activity of the most common typologies used in hospitals. At current, a dataset for the following distinct classes has been collected: gauzes, latex gloves (single and pair), nitrile gloves (single and pair), surgery gloves (single and pair), medical caps, medical glasses, shoe covers (single and pair), test tubes, and urine bags. Sample images are reported in Figure \ref{fig:dataset}. Over 1400 image triplets were already published in the public dataset \cite{dataset}, while a more structured dataset with over 2100 image triplets will be made available in the near future.

\subsection{Model definition and training}
\label{sec:cnn}
Since the image in the dataset have a standard background and most of the image area is occupied by an object of interest, the computer vision task to be solved has been represented as an image classification pipeline.  In the future, a detector of the area of interest might be included to exclude areas irrelevant to the classification. Similarly, an object detection method might be  used to replace the simple image classification, thus providing g also information on the location of the object of interest in the scene.
Based on our previous experience \cite{bruno2022efficient,bruno2022exploring}, it has been evaluated to assess the performance of convolutional neural networks belonging to the EfficientNet family \cite{tan2019EfficientNet}. As the name suggests, EfficientNet improves the classification quality with lower complexity compared to models having similar classification performances. This is possible since EfficientNet performs optimised network scaling, given a predefined complexity. Eight EfficientNet networks named b0-b7 are available with different levels of complexity. Since we aimed at a sustainable and on-the-edge analysis of the images in the place where primary sorting is done, the less complex network b0 has been considered, having only 5.5M parameters with respect to b7 coming with 66M parameters. For the time being, 3D information has not been included and some of the classes were merged for the classification task. Namely, the six classes in the dataset corresponding to gloves of different materials and as single items or in pairs were merged. Similarly, single items and pairs of shoe covers were merged, resulting in a total of 7 macro-classes. Therefore, the final fully connected layers of EfficientNet b0 were changed to accommodate the desired output size.

\section{Experimental results}
\label{sec:exp}
\begin{figure}
    \centering
   
\includegraphics[width=0.46\textwidth]{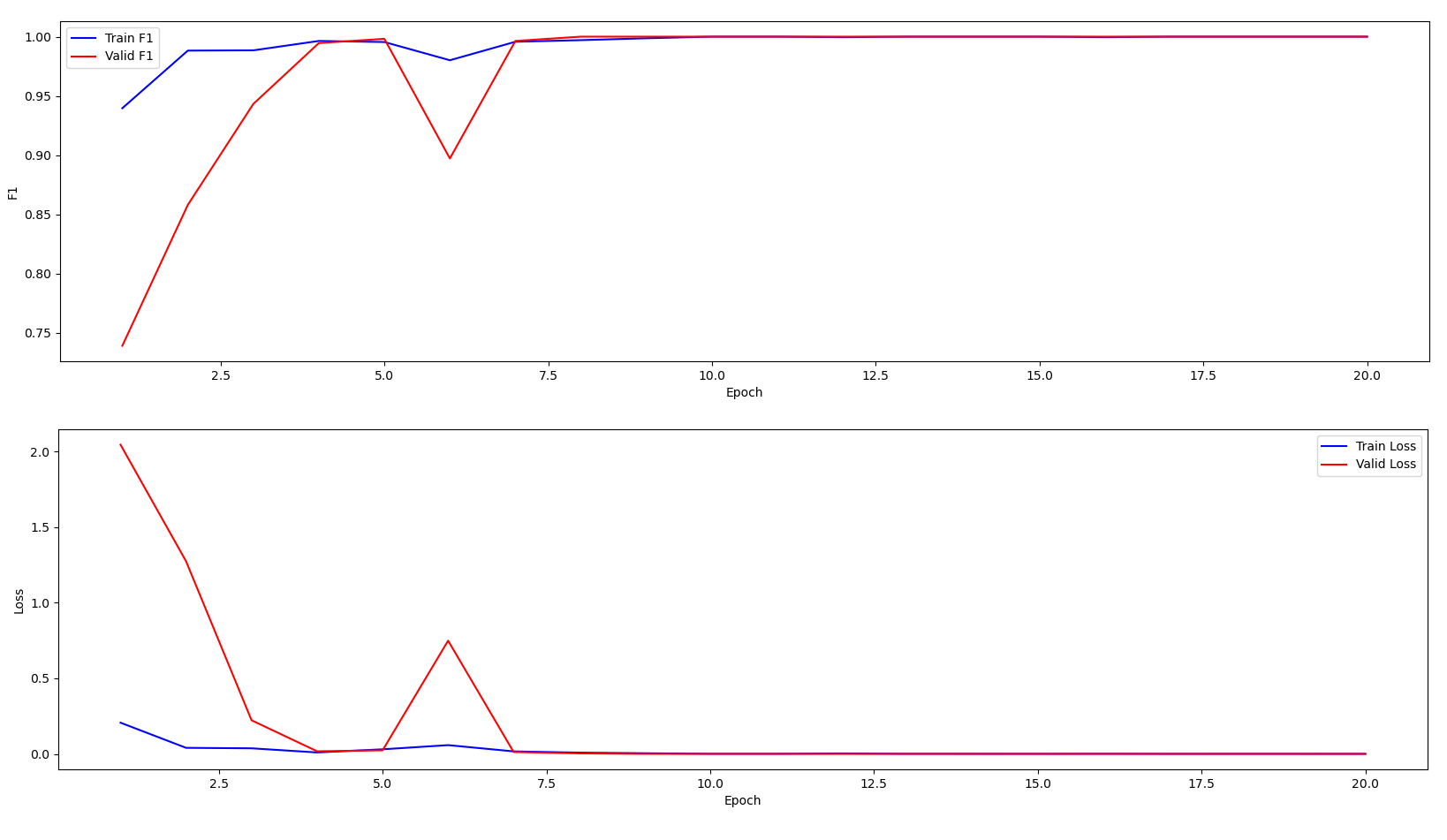}
    \caption{Training and validation loss on top and training and validation accuracy curves at bottom for model training on dataset $\mathcal{A}$}
    \label{fig:train}
\end{figure}

A version of EfficientNet b0 pretrained on the COCO Dataset has been used in the experiments adapting the final layer to fit the classification in the 7 classes described before. The initial dataset, named dataset $\mathcal{A}$, has been split into the train (4416 images), validation (551 images) and test (553 images) sets.  Images were not preprocessed in any way; no augmentation  has been used in the present paper. The training and validation losses and the relative accuracies plots (Figure \ref{fig:train}) show that in 10 steps the maximum result is obtained, and an early stopping set to ten avoids overfitting. After training, the network was used to perform inference on the test set: no error in classification was observed achieving a 100\% accuracy. An additional test was executed by acquiring new data and by building a new dataset named $\mathcal{B}$. The dataset $\mathcal{B}$ was acquired  on different days from the dataset $\mathcal{A}$, in order to avoid any possible duplicated images between datasets $\mathcal{A}$ and $\mathcal{B}$. The results of inference on dataset $\mathcal{B}$ are reported in Table \ref{tab:perf}, showing an accuracy of 99.54\%, demonstrating the generalisation capabilities of the trained model.


\begin{table}[h]
\caption{Performance of the model trained on dataset $\mathcal{A}$ during testing in the dataset $\mathcal{B}$.}\label{tab:perf}
\newcommand\T{\rule{0pt}{2.5ex}}
\begin{center}
\begin{tabular}{lcccc}
 {\bf Class} &  {\bf Precision} & {\bf Recall} &  {\bf F1-Score} &  {\bf Support}\\
\hline
\hline
\texttt{Gauzes} &  0.98913 &  0.97849  & 0.98378   &     93 \\
\texttt{Gloves}  &  0.99209 &  0.99603  & 0.99406   &    252 \\
\texttt{Med Cap} &    1.00000 &  1.00000  & 1.00000  &      57 \\
\texttt{Med Glasses} &   1.00000 &  1.00000  & 1.00000 &       93 \\
\texttt{Shoe Cover} &   0.99265 &  1.00000  & 0.99631   &    135 \\
\texttt{Test Tube} &    1.00000  & 1.00000 &  1.00000   &     54 \\
\texttt{Urine Bag} &    1.00000 &  0.97778  & 0.98876   &     45 \\
\hline
Accuracy  &&&                      0.99451    &   729\\
Macro avg &   0.99627 &  0.99319  & 0.99470   &    729\\
Weighted avg  &  0.99452 &  0.99451  & 0.99450  &     729
 \\\hline\hline
\end{tabular}
\end{center}

\label{tab:comparisons}
\end{table}
 
\section{Conclusions and future works}
\label{sec:conclusion}
In this paper, we have introduced the importance of primary sorting in medical waste management and contributed to the definition of approaches for assisting operators in such a task. To this end, we introduced a computer vision system based on deep learning. The approach was tested on datasets we collected ad hoc.
In the future, we plan to extend the datasets so that they can serve as a valuable resource for the research community and provide a benchmark for evaluating medical waste sorting algorithms using computer vision. To make our dataset widely accessible, we have decided to release it and make it available for research purposes publicly. We hope this contribution will encourage further research in the field and help advance the development of medical waste sorting using computer vision.

\section*{Acknowledgement}
\label{sec:ack}
This research was partially supported by the Medical Waste Treating 4.0 project funded by the Tuscany Region, Italy.

\bibliographystyle{IEEEbib}
\bibliography{strings,refs}

\end{document}